%% file: 00-root.tex
\documentclass[sigconf]{acmart}

\usepackage{booktabs} 
\usepackage{makecell}
\usepackage{tcolorbox}
\usepackage{subcaption}
\usepackage{tablefootnote}
\usepackage{arydshln}
\usepackage{colortbl}
\usepackage{xcolor}
\usepackage{xspace}
\usepackage{multirow}
\usepackage{tabularx}
\usepackage{graphicx}
\usepackage[inline]{enumitem}

\usepackage{XX-slnotation}

\AtBeginDocument{%
  }

\acmYear{2026}
\setcopyright{cc}
\setcctype{by}

\acmConference[SIGIR '26] {Proceedings of the 49th International ACM SIGIR Conference on Research and Development in Information Retrieval}{July 20--24, 2026}{Melbourne, VIC, Australia.}
\acmBooktitle{Proceedings of the 49th International ACM SIGIR Conference on Research and Development in Information Retrieval (SIGIR '26), July 20--24, 2026, Melbourne, VIC, Australia}
\acmISBN{979-8-4007-2599-9/2026/07}
\acmDOI{10.1145/3805712.3809954}



\begin{document}

\title{LTRR: Learning To Rank Retrievers for LLMs}

\author{To Eun Kim}
\orcid{0000-0002-2807-1623}
\affiliation{%
  \institution{Carnegie Mellon University}
  \city{Pittsburgh}
  \state{PA}
  \country{USA}
}
\email{toeunk@cs.cmu.edu}

\author{Fernando Diaz}
\orcid{0000-0003-2345-1288}
\affiliation{%
  \institution{Carnegie Mellon University}
  \city{Pittsburgh}
  \state{PA}
  \country{USA}
}
\email{diazf@acm.org}

\input{XX-notation}

\input{00-abstract}
\maketitle
\input{01-intro}

\input{02-related-work}

\input{03-rag-pipeline}

\input{03-method}

\input{04-exp}

\input{05-results}

\input{06-conclusion}

\begin{acks}
This work was supported by the National Science Foundation under Grant No. 2402874.
Any opinions, findings and conclusions or recommendations expressed in this material are those of the authors and do not necessarily reflect those of the sponsors.
\end{acks}

\bibliographystyle{ACM-Reference-Format}
\bibliography{XX-references.bib}

\newpage
\appendix
\input{99-appendix}

\end{document}

%% file: XX-notation.tex


\renewcommand{\slReals}{\mathbb{R}}


\let\inputSpace\mlInputs
\let\instance\mlInput
\let\outputSpace\mlOutputs
\renewcommand{\mlOutput}{\hat{y}}
\let\output\mlOutput
\renewcommand{\mlTarget}{y}
\let\target\mlTarget

\newcommand{\query}{q}
\newcommand{\querySpace}{\mathcal{Q}}

\newcommand{\rModel}{\mathcal{R}}

\newcommand{\ltrrScorer}{\mlPredictor}
\newcommand{\ltrrRouter}{\mathcal{F}}

\newcommand{\features}{\Phi}
\newcommand{\preRetrievalFeatures}{\Phi_{pre}}
\newcommand{\postRetrievalFeatures}{\Phi_{post}}

\newcommand{\llmModel}{\mathcal{G}}
\newcommand{\queryGenerator}{\phi_q}
\newcommand{\promptGenerator}{\phi_p}
\newcommand{\augmentedPrompt}{\overline{\instance}}

\newcommand{\doc}{d}
\newcommand{\corpus}{\mathcal{D}}

\newcommand{\retrievalResult}{z}


\newcommand{\utilityGain}{\delta}
\newcommand{\utilityMetric}{\evalMetric_u}
\newcommand{\precisionMetric}{\evalMetric_p}
\newcommand{\recallMetric}{\evalMetric_r}
\newcommand{\faithfulnessMetric}{\evalMetric_f}

\newcommand{\efivebase}{$\text{E5}_{base}$\xspace}

\newcommand{\xgboost}{XGBoost\xspace}
\newcommand{\svmrank}{SVMRank\xspace}
\newcommand{\ffn}{FFN\xspace}
\newcommand{\listnet}{ListNet\xspace}
\newcommand{\lambdamart}{LambdaMart\xspace}
\newcommand{\deberta}{DeBERTa\xspace}

\newcommand{\overallsim}{\texttt{OverallSim}\xspace}
\newcommand{\avgsim}{\texttt{AvgSim}\xspace}
\newcommand{\maxsim}{\texttt{MaxSim}\xspace}
\newcommand{\varsim}{\texttt{VarSim}\xspace}
\newcommand{\moran}{\texttt{Moran}\xspace}

\newcommand{\tekcomment}[1]{\textcolor{blue}{[TEK: #1]}}
\newcommand{\fdcomment}[1]{\textcolor{red}{[FD: #1]}}

%% file: 00-abstract.tex
\begin{abstract}
Retrieval-Augmented Generation (RAG) systems typically rely on a single fixed retriever, despite growing evidence that no single retriever performs optimally across all query types. In this paper, we explore a query routing approach that dynamically selects from a pool of retrievers based on the query, using both train-free heuristics and learned routing models. We frame routing as a learning-to-rank problem and introduce LTRR, a framework that Learns To Rank Retrievers according to their expected contribution to downstream RAG performance. Through experiments on diverse question-answering benchmarks with controlled variations in query types, we demonstrate that routing-based RAG consistently surpasses the strongest single-retriever baselines. The gains are particularly substantial when training with the Answer Correctness (AC) objective and when using pairwise ranking methods, with XGBoost yielding the best results. Additionally, our approach exhibits stronger generalization to out-of-distribution queries. Overall, our results underscore the critical role of both training strategy and optimization metric choice in effective query routing for RAG systems.
\end{abstract}

%% file: 01-intro.tex
\section{Introduction}\label{sec:intro}

\begin{figure}[t]
    \centering
    \resizebox{0.9\columnwidth}{!}{ 
        \includegraphics[trim=200 80 160 95, clip]{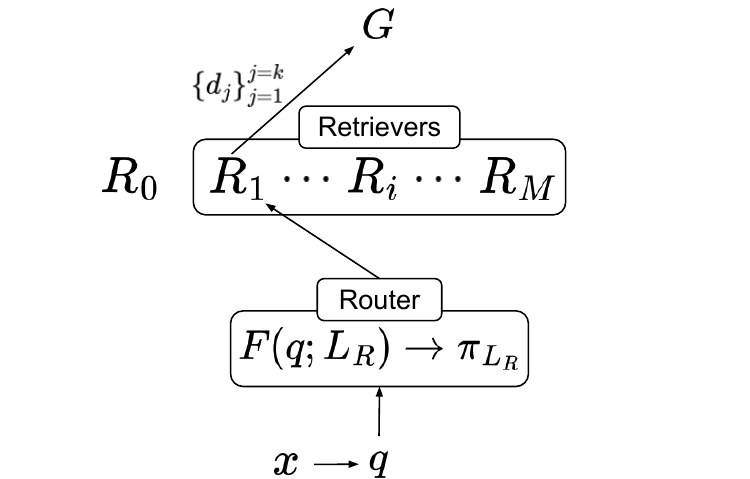}
    }
    \caption{Abstracted query routing workflow in our system. An input question $\instance$ is formulated into a query, which is passed to a router $\ltrrRouter$ that produces a ranking over available retrievers. The top-ranked retriever is selected and its document set is passed to generator $\llmModel$. The router is optimized to rank retrievers by their expected downstream RAG utility as judged by the generator. Crucially, our setup includes a no-retrieval option $\rModel_0$, allowing the router to jointly address the \textit{when-to-retrieve} and \textit{where-to-retrieve} problems during selection. See \S\ref{sec:rag} and \S\ref{sec:ltrr} for full details.}
    \label{fig:router-pipeline}
    \vspace{-10pt}
\end{figure}

Retrieval-Augmented Generation (RAG) \cite{lewis2020retrieval, kim2024reml} enhances Large Language Models (LLMs) by leveraging external knowledge. 
However, despite the availability of numerous retrieval models, no single retriever consistently outperforms others across diverse queries and corpora~\cite{Khramtsova23whichRetriever, wang2022text}. 
Inspired by distributed information retrieval \cite{callan2002distributed} and federated search \cite{diaz2010federatedAggregated}, the research community in retrieval-enhanced machine learning \cite{kim2024reml, zamani:reml} has recently begun investigating \textit{query routing} as a means to address this challenge.

Effective query routing is essential given the complexity introduced by various retrieval methods. Optimal routing can improve adaptability across query types, thereby enhancing downstream RAG performance \cite{Khramtsova24RetrieverRanking}. Additionally, selective routing can help reduce computational costs by activating only necessary retrievers, including potentially bypassing retrieval entirely. Furthermore, modern information access systems increasingly operate as competing services, effectively forming a \textit{search marketplace} targeted at machine consumers (LLMs)~\cite{luo2025evaluation}, making efficient query routing a timely research.

Existing query routing approaches exhibit several limitations. Many rely on heuristics~\cite{Khramtsova23whichRetriever, lee2024routerretriever} or optimize solely for traditional retrieval metrics intended for human end-users~\cite{Guerraoui25FederatedRAG}, while in the context of LLM consumers, routing strategies should ideally optimize downstream LLM performance rather than traditional retrieval metrics \cite{kim2025fairragimpactfair,salemi2024erag}.
Additionally, these methods often leverage query-corpus similarity-based routing~\cite{Khramtsova23whichRetriever, Khramtsova24RetrieverRanking, lee2024routerretriever, Guerraoui25FederatedRAG}, which becomes impractical in \textit{uncooperative} environments~\cite{arguello2017aggregated}, where we do not have access to the corpus, and the retrievers are assumed to only provide a search interface (e.g., Model Context Protocol \cite{anthropic2024mcp}).

In this paper, we propose \textbf{Learning to Rank Retrievers (LTRR)}, a novel query routing approach explicitly optimized for downstream LLM utility. Unlike previous work, our ranking of retrievers is based directly on the relative improvement in the LLM's generation quality compared to a no-retrieval baseline. This approach inherently addresses both retriever-selection (``where to query") \cite{Khramtsova23whichRetriever} and retrieval-necessity (``when to query") \cite{mallen-etal-2023-trust}, as \textit{no-retrieval} is explicitly included as one of the routing options.
We comprehensively evaluate multiple learning-to-rank (LTR) algorithms using diverse query sets, focusing on retrievers differentiated by retrieval strategies rather than corpus content. This setting reflects contemporary retrieval environments, where systems typically utilize similar large-scale corpora but differ significantly in their retrieval approaches.

Our experiments demonstrate that LTRR algorithms, particularly those trained using pairwise \xgboost, significantly outperform the best standard single-retriever-based RAG systems in both in-distribution and some out-of-distribution evaluations.
We release our code for reproducibility and to encourage further research in this direction.\footnote{\url{https://github.com/kimdanny/Starlight-LiveRAG}}

%% file: 02-related-work.tex
\section{Related Work}\label{sec:related-work}

\paragraph{\textbf{Distributed IR}}
Our approach to query routing builds on several established IR fields, including Distributed Search, Federated Search, Selective Search, Aggregated Search, and Meta Search. Distributed search traditionally address the selection of relevant document collections based on query relevance, often in a distributed and disjoint environment~\cite{callan1995searching, callan2002distributed}. Federated and selective search extend these ideas, focusing on brokered retrieval across multiple independent and typically uncooperative systems, employing resource representations and selection strategies to effectively route queries~\cite{diaz2010federatedAggregated, dai2017ltrResources}. Aggregated Search similarly aims to integrate diverse retrieval results from specialized vertical search services into a unified search interface, emphasizing the selection of relevant services per query~\cite{arguello2017aggregated}. Additionally, Meta Search combines results from several search engines to improve overall relevance, recognizing that no single search engine consistently outperforms others across diverse queries~\cite{glover1999metasearch, chen2001metaspider}.

While our query routing methodology shares conceptual similarities with these fields, it uniquely differs in its explicit emphasis on routing queries to varied retrieval strategies optimized directly for downstream retrieval-augmented generation (RAG) tasks.

\paragraph{\textbf{Query Routing Strategies}} 
Building on insights from distributed IR, recent RAG systems increasingly incorporate query routing strategies. \citet{Khramtsova23whichRetriever} examined various dense retriever selection methods and identified corpus similarity as a reliable, training-free criterion. This work was extended by \citet{Khramtsova24RetrieverRanking}, who proposed an unsupervised approach using pseudo-query generation and LLM-based pseudo relevance judgments to rank dense retrievers.

\citet{mu-etal-2024-query} proposed a routing strategy that directly predicts downstream LLM performance, bypassing traditional retrieval effectiveness metrics. However, this approach overlooks cases where retrieval may not be beneficial and struggles with the variability of absolute score prediction across queries. Similarly, Adaptive-RAG \cite{jeong-etal-2024-adaptiveRAG} classifies queries by their perceived complexity to select retrieval strategies, but this relies on human-centric definitions of complexity and requires curated training data, which may not align with LLM behavior.
Other recent studies expand the space of query routing. RouterRetriever~\cite{lee2024routerretriever} uses embedding similarities for retriever selection. \citet{Guerraoui25FederatedRAG} introduced RAGRoute, a lightweight classifier that dynamically selects retrieval sources to optimize recall and classification accuracy. \citet{tang2024adapting} framed routing as a contextual multi-armed bandit problem in knowledge graph-based RAG systems, but without modeling no-retrieval as a viable option. Complementary to selection-based routing, \citet{kalra-etal-2025-mor} proposed MoR, which instead aggregates all available retrievers via weighted fusion, forgoing explicit retriever selection in favor of combining complementary strengths.

Our approach emphasizes learning to rank retrievers based directly on improvements in downstream LLM utility. It explicitly includes no-retrieval as a valid action and is evaluated over a diverse set of retrieval strategies.

%% file: 03-rag-pipeline.tex
\section{RAG Method with Query Routing}\label{sec:rag}

\paragraph{\textbf{Multi-Retriever Setup}}
We utilize BM25 \cite{robertson2009probabilistic} and \efivebase \cite{wang2022text} dense retrievers as a base retrievers, and we combine two reranking strategies with distinct goals to make variations in retrieval strategies. The first, \textit{score regularization}~\cite{diaz2007regularizing-querybased}, focuses on improving retrieval performance. The second, \textit{stochastic reranking}~\cite{kim2025fairragimpactfair}, aims to enhance item-fairness and diversity, which can also improve downstream RAG performance. As a result, we establish six distinct retrievers:
(1) BM25; 
(2) BM25 + Score Regularization Reranking;
(3) BM25 + Stochastic Reranking;
(4) \efivebase;
(5) \efivebase + Score Regularization Reranking; and
(6) \efivebase + Stochastic Reranking.
Details of reranking methods can be found in Appendix \ref{sec:app:reranking}.

All retrievers utilize the sampled version of FineWeb corpus (15M documents)\cite{penedo2025fineweb}, and their retrieval strategies and corpus statistics remain hidden from both the generator and router (uncooperative environment).

\paragraph{\textbf{Query Routing via LTRR}}
Our RAG framework routes queries to a suitable retriever $\rModel_i$ from a pool of multiple retrievers $L_\rModel$. For each input instance $\instance$, we generate a query $\query$ via a query generation function $\queryGenerator(\instance)$. The core objective is to route this query to one or more retrievers that maximize the downstream performance of the RAG system.
Formally, we introduce a router function $\ltrrRouter$ that maps queries to a ranked set of retrievers:
\vspace{-2pt}
\begin{align}
    \ltrrRouter(\query;L_\rModel)\rightarrow \raRanking_{L_\rModel} ,
\end{align}

where $\raRanking_{L_\rModel}$ is a ranking of retrievers reflecting the predicted utility of each retriever can give to the downstream generator $\llmModel$ for the given query.
In our implementation, we route queries to the top-ranked retriever using a pairwise \xgboost-based router (chosen for its empirical effectiveness, as discussed in later sections). Importantly, we include a `no-retrieval' option $\rModel_0$ in the ranking. This allows the system to bypass retrieval altogether when the router predicts that relying solely on the language model's parametric memory yields the best performance.

\paragraph{\textbf{Generator}}
We use Falcon3-10B-Instruct~\cite{Falcon3} as the generator in our RAG system. Inspired by recent work on prompting LLMs to reason with external information~\cite{jin2025searchR1, chen2025ReSearch}, we design prompts that instruct the model to explicitly assess the relevance and utility of each retrieved passage. Specifically, the model is prompted to reflect on how to use the passages in a \texttt{<think>} section, followed by its final answer in a \texttt{<answer>} section. We extract only the content within the \texttt{<answer>} tag as the system's output. For ill-formatted generations, a fallback prompt omitting explicit reasoning is used. Full prompt details are provided in the codebase and Appendix \ref{sec:app:prompts}.

%% file: 03-method.tex
\section{Learning to Rank Retrievers}\label{sec:ltrr}

We propose Learning to Rank Retrievers (LTRR), which formulates the routing problem as a learning-to-rank task \cite{liu2009learning} tailored specifically to optimize downstream LLM performance.

To first derive the ground-truth retriever rankings required for training, we measure the utility gain $\utilityGain_i$ achieved by retriever $\rModel_i$ relative to the baseline generator performance (without retrieval):
\begin{align}
    \utilityGain_i = \utilityMetric(\llmModel(\augmentedPrompt_i), \target) - \utilityMetric(\llmModel(\instance), \target)  ,
\end{align}
where $\augmentedPrompt_i = \promptGenerator(\instance, \rModel_i(\query, k))$ with $k$ denoting the number of passages to retrieve, $\promptGenerator$ denoting a prompt construction function for an LLM $\llmModel$, and $\utilityMetric$ is an arbitrary string utility metric. To ensure comparability across queries, utility-gain scores are min-max normalized per query into the range of [0,1].

Following the LTR literature \cite{cao2007ltr}, LTRR is then characterized by a scoring function $\ltrrScorer(\features(\query, \rModel_i)) \rightarrow \slReals$
that assigns a score to each retriever based on query- and retriever-specific features $\features(\query, \rModel_i)$ extracted from the $i$'th retriever $\rModel_i$.

To train the ranking model, we experiment with three well-established approaches (detailed in Appendix \ref{sec:app:ltrr-loss}): (1) the \textit{pointwise} approach, where the model predicts each retriever's utility gain $\utilityGain_i$ independently using a regression loss; (2) the \textit{pairwise} approach, where the model learns to minimize ranking inversions between retriever pairs based on their relative utility gains compared to the no-retrieval baseline; and (3) the \textit{listwise} approach, which directly optimizes the predicted utility gains over the full set of retrievers for each query.

\subsection{LTR Features}
Our setup assumes an \textit{uncooperative} retrieval environment in which retrievers do not expose detailed corpus statistics or embedding model specifications. Thus, we extract a set of query-dependent pre-retrieval features and query- and retriever-dependent post-retrieval features to facilitate effective learning-to-rank (LTR) modeling.

For \textbf{pre-retrieval features}, we include the query representation ($\slReals^{\text{dim}}$), query length, and query type. The query representation is a vector produced by an embedding model, with optional dimensionality reduction (e.g., via PCA).\footnote{\efivebase model was used to extract query embeddings, which are then reduced to 32 dimensions using PCA. For \deberta-based models, however, we retain the original, non-reduced embeddings.} Query type is determined using a trained lightweight classifier that distinguishes between keyword-based and natural language queries.
These features are query-specific but not retriever-specific, allowing LTRR models to learn differences across queries.

\textbf{Post-retrieval features}, in contrast, are computed after querying all retrievers and are both query- and retriever-specific, providing the LTRR model with signals to differentiate between retrievers.

Let $\retrievalResult_i = [\doc_{i,1}, \dots, \doc_{i,k}]$ denote the top-k documents retrieved by retriever $\rModel_i$, $s(\cdot,\cdot)$ be an embedding-based cosine similarity function, and $M$ be the total number of available retrievers. We define $e(\retrievalResult_i) = \frac{1}{k}\sum^{k}_{j=1} \texttt{embed}(\doc_{i,j})$ as the aggregated semantic embedding of the retrieved documents retrieved by $\rModel_i$.
Using these definitions, we construct the following semantic and statistical features:
\begin{itemize}[leftmargin=15pt]
\item \overallsim: similarity between query and the aggregated embedding of retrieved documents, $s(\query, e(\retrievalResult_i))$,
\item \avgsim: average similarity score between query and individual retrieved documents, $\texttt{avg}_j(s(\query, \doc_{i,j}))$,
\item \maxsim: maximum similarity score between query and individual retrieved documents, $\texttt{max}_j(s(\query, \doc_{i,j}))$,
\item \varsim: variance of retrieval similarity scores, $\texttt{var}_j(s(\query, \doc_{i,j}))$, capturing retrieval confidence dispersion,
\item \moran: Moran coefficient~\cite{diaz2007performance}, which measures semantic autocorrelation among retrieved documents in alignment with the cluster hypothesis, and
\item \texttt{CrossRetSim}: average semantic similarity of the current retriever's result set with those from other retrievers, defined as $\frac{1}{M-1} \sum^{M}_{m=1; m\ne i} s(e(\retrievalResult_i), e(\retrievalResult_m))$, which can indicate a uniqueness of a ranking compared to other rankings from other retrievers.
\end{itemize}

For the no-retrieval option ($\rModel_0$), only pre-retrieval features are available. To maintain consistent feature dimensionality across retrievers, we handle missing post-retrieval features differently depending on the model type. For neural LTR models, we introduce a dedicated learnable parameter vector to represent the post-retrieval feature space of $\rModel_0$. This vector is randomly initialized and optimized during training, allowing the model to implicitly encode the characteristics of the no-retrieval strategy. For non-neural models, we apply median imputation based on the training data to fill in the missing post-retrieval features, ensuring compatibility with fixed-length feature inputs.

%% file: 04-exp.tex
\section{Experiment}\label{sec:exp}
We evaluate our proposed routing approaches against a range of baseline and train-free routing models.

\subsection{Routing Models}
We first consider five heuristic, train-free routing models, each based on a post-retrieval feature: \overallsim, \avgsim, \maxsim, \varsim (where lower variance is preferred), and \moran.

For learned routing models trained via the LTRR framework, we evaluate eleven models spanning the \textbf{pointwise}, \textbf{pairwise}, and \textbf{listwise} paradigms. In the pointwise and pairwise settings, we train models using \xgboost~\cite{chen2016xgboost}, \svmrank~\cite{joachims2006training}, a feedforward network (\ffn), and \deberta~\cite{he2021debertav3}. In the listwise setting, we evaluate \listnet~\cite{cao2007ltr}, \lambdamart~\cite{wu2010adapting}, and \deberta-based models.

All LTRR models are trained using utility labels derived from two metrics: BEM~\cite{bulian2022-tomayto-bem} and Answer Correctness (AC)~\cite{es-etal-2024-ragas}, both shown to correlate strongly with human evaluations in RAG setting \cite{oro2024evaluating}.

\input{05-results/table}

\subsection{Datasets}
We require controlled and diverse variations in query types in order to evaluate routing behavior under heterogeneous information needs.
In particular, since routing decisions are intended to adapt to differences in question characteristics and user profiles, the evaluation dataset must expose systematic variation along these dimensions.

To this end, we generate a synthetic dataset using DataMorgana~\cite{filice2025datamorgana}, which provides fine-grained control over both question and user attributes.
Question configurations span four dimensions: answer type, premise, phrasing, and linguistic variation, while user configurations capture differences in expertise level (expert vs. novice).
We follow \citet{filice2025datamorgana} for the detailed dataset generation procedure. Full dataset generation details are provided in Appendix \ref{sec:app:dm-tables}.

For the LTRR experiments, we focus on the answer-type category, which comprises five distinct question types: factoid, multi-aspect, comparison, complex, and open-ended.

We construct five dataset splits for evaluation. The \textbf{Balanced split} includes all question types proportionally in both training and test sets. Four \textbf{unseen type splits} (multi-aspect, comparison, complex, and open-ended) each hold out one question type from training and use it exclusively for testing, enabling us to assess model generalization to unseen query types. 
Dataset statistics are reported in Appendix~\ref{sec:app:data-stat}.

%% file: 05-results/table.tex
\begin{table*}[]
\centering
\small\caption{Average downstream RAG utility measured by either BEM or AC when the test queries are routed to the top retriever based on the routing model. \textbf{Bold}: statistically significant improvement over the best standard RAG system according to the paired Wilcoxon signed-rank tests with bonferroni correction.
}
\resizebox{\textwidth}{!}{
\begin{tabular}{c ccccc c ccccc}
   & \multicolumn{5}{c}{BEM-based}   && \multicolumn{5}{c}{AC-based} \\
   \cline{2-6} \cline{8-12}
   Model & \makecell{Balanced} & \makecell{Unseen\\multi-aspect}   & \makecell{Unseen\\comparison} & \makecell{Unseen\\complex} & \makecell{Unseen\\open-ended} && \makecell{Balanced} & \makecell{Unseen\\multi-aspect}   & \makecell{Unseen\\comparison} & \makecell{Unseen\\complex} & \makecell{Unseen\\open-ended} \\
   \hline
Oracle-Router & 0.5753 & 0.3401 & 0.5217  & 0.5799    & 0.5601  && 0.7948 & 0.7802 & 0.7831 & 0.7978 & 0.7852 \\
\cline{1-6} \cline{8-12}
Best Standard RAG & 0.3599 & 0.1941 & 0.3074  & 0.3595  & 0.3465  && 0.5776 & 0.5809 & 0.5723 & 0.5812 & 0.5784 \\
\cline{1-6} \cline{8-12}
\overallsim-Router & 0.3474 & 0.1874 & 0.3004  & 0.3458    & 0.3261  && 0.5722 & 0.5835 & 0.5689 & 0.5668 & 0.5658 \\
\avgsim-Router & 0.3542 & 0.1855 & 0.2995  & 0.3454    & 0.3438  && 0.5855 & 0.5871 & 0.5724 & 0.5825 & 0.5739 \\
\maxsim-Router & 0.3623 & 0.1913 & 0.3161 & 0.3591    & 0.3521  && 0.5872 & 0.5777 & 0.5753 & 0.5891 & 0.5775 \\
\varsim-Router & 0.3307 & 0.1864 & 0.3082  & 0.3255 & 0.2941  && 0.5593 & 0.5811 & 0.5553 & 0.5603 & 0.5450 \\
\moran-Router & 0.3304 & 0.1880 & 0.3083  & 0.3255    & 0.2941  && 0.5603 & 0.5840 & 0.5535 & 0.5596 & 0.5462 \\
\cline{1-6} \cline{8-12}
$XGBoost_{point}$ & 0.3444 & 0.1915 & 0.3065  & 0.3519    & 0.3318  && 0.5760 & 0.5925 & 0.5698 & 0.5790 & 0.5885 \\
$SVMRank_{point}$ & 0.3548 & 0.1894 & 0.3167  & 0.3624    & 0.3284  && \textbf{0.5903} & 0.5815 & 0.5779 & 0.5899 & 0.5885\\
$FFN_{point}$ & 0.3528 & 0.1894 & 0.3039  & 0.3397    & 0.3211  && 0.5760 & 0.5901 & 0.5650 & 0.5854 & 0.5529 \\
${DeBERTa}_{point}$ & 0.3604 & 0.1911 & 0.3064  & 0.3590  & 0.3387  && \textbf{0.5884} & 0.5847 & 0.5762 & 0.5925 & 0.5863 \\
\cdashline{1-6} \cdashline{8-12}
$XGBoost_{pair}$ & 0.3626 & 0.1909 & 0.3094  & 0.3625   & 0.3504  && \textbf{0.5900} & 0.5769 & 0.5759 & \textbf{0.6017} & \textbf{0.5930} \\
$SVMRank_{pair}$ & 0.3657 & 0.1883 & 0.2943  & 0.3643   & 0.3605  && 0.5329 & 0.5842 & 0.5695 & 0.5377 & 0.5791 \\
$FFN_{pair}$ & 0.3575 & 0.1892 & 0.3015  & 0.3654   & 0.3457  && 0.5831 & 0.5872 & 0.5814 & 0.5798 & 0.5869 \\
${DeBERTa}_{pair}$ & 0.3634 & 0.1956 & 0.3083  & 0.3654   & 0.3445  && \textbf{0.5888} & 0.5805 & 0.5777 & 0.5912 & 0.5851 \\
\cdashline{1-6} \cdashline{8-12}
$ListNet_{list}$ & 0.3530 & 0.1934 & 0.3032  & 0.3398    & 0.3254  && 0.5848 & 0.5836 & 0.5718 & 0.5847 & 0.5818 \\
$LambdaMART_{list}$ & 0.3450 & 0.1959 & 0.2934  & 0.3631 & 0.3229  && 0.5802 & 0.5812 & 0.5483 & 0.5690 & 0.5662 \\
${DeBERTa}_{list}$ & 0.3235 & 0.1825 & 0.3005  & 0.3470  & 0.2736  && 0.5829 & 0.5902 & 0.5556 & 0.5434 & 0.5751 \\
\end{tabular}
}
\label{tab:main-results}
\vspace{-10pt}
\end{table*}

%% file: 05-results.tex
\section{Results and Discussion}\label{sec:results}

\paragraph{\textbf{RQ1: Do routing-based RAG systems outperform the best-performing standard RAG model?}}

To study this question, we first identify the best-performing single-retriever (standard) RAG system for each dataset under the two utility metrics. Their downstream scores are shown in the `Best Standard RAG' row of Table~\ref{tab:main-results}.
See Appendix~\ref{sec:app:single-ret} for details.

As shown in Table~\ref{tab:main-results}, the train-free routing models did not yield statistically significant improvements over the best-performing standard RAG systems, despite showing some numerical gains. In contrast, the LTRR-based models demonstrated more substantial improvements, particularly with the pointwise \svmrank and \deberta, as well as the pairwise \xgboost and \deberta models, which achieved statistically significant gains on the Balanced split.

However, performance gains were noticeably higher for router models trained using the AC utility metric compared to those trained with BEM. Statistically significant improvements were observed only for AC-based routers (highlighted in bold), while no such gains were found for BEM-based models. We attribute this discrepancy to differences in metric reliability: although both BEM and AC correlate well with human judgments, prior work shows that AC consistently achieves stronger alignment \cite{oro2024evaluating}. Since LTRR models are trained directly on utility labels, the choice of a consistent and accurate metric is critical.

\paragraph{\textbf{RQ2: Do LTRR-based routing algorithms outperform the best-performing train-free routing model?}}

We also examined whether trained routing models (LTRR-based) outperform the highest train-free baselines. As in RQ1, numerical results indicate that LTRR models generally outperform the strongest train-free routers (usually \maxsim). However, statistical significance tests revealed that these improvements were not significant after correction. This suggests that the observed gains may be subject to variability and underscores the need for larger-scale studies or refined methods to more conclusively demonstrate the advantages of trained routing over train-free approaches.

\paragraph{\textbf{RQ3: Are the performance improvements from routing-based RAG models robust across the different unseen query-type splits?}}

We investigated the robustness of performance improvements across various unseen query types (multi-aspect, comparison, complex, open-ended). While train-free routing models showed relatively modest improvements across these splits, LTRR-based trained routing algorithms displayed more stable and consistent performance gains. In particular, the pairwise \xgboost routers trained on the AC utility metric showed the most consistent outperformance over the standard RAG and train-free baselines across different unseen datasets, achieving statistically significant results in the complex and open-ended query splits.

\paragraph{\textbf{Discussion and Implications}}
Our findings underscore that not all routing improvements are created equal. While LTRR models often outperform standard and train-free routing methods numerically, only a subset achieve statistically significant gains, particularly those trained with the AC utility metric. This confirms that metric choice is not merely a technical detail, but a determinant of learning signal quality.
Moreover, the effectiveness of pairwise training (especially with tree-based models like \xgboost) suggests that explicitly modeling retriever tradeoffs per query offers a more robust inductive bias than listwise or pointwise formulations. The observed performance robustness of LTRR across unseen question types further indicates that routing functions can generalize beyond their training distribution. This points to the potential of query routing as a critical component in adaptive retrieval architectures, especially for long-tailed or evolving query scenarios.
Notably, our system is built entirely on lightweight, cost-effective retrievers and a computationally efficient routing model, but still achieves meaningful gains, showing that even modest setups can benefit from query routing.
Finally, since LTRR produces a full ranking over retrievers, it naturally supports future extensions to multi-retriever selection, where retrieved results can be fused to enhance coverage and diversity \cite{cormack2009reciprocal}. We leave this extension for future work.

%% file: 06-conclusion.tex
\section{Conclusion}\label{sec:conclusion}

We introduce LTRR, a query routing framework that learns to rank retrievers based on their downstream utility to large language models. Our empirical results show that RAG systems equipped with trained routers---particularly those using the AC utility metric and pairwise learning-to-rank algorithms such as \xgboost---can outperform standard single-retriever RAG systems and generalize to unseen query types. These findings highlight the importance of utility-aware retriever selection and demonstrate that learning-based query routing offers a promising path toward more robust and adaptable RAG systems.

%% file: 99-appendix.tex
\onecolumn

\section{Reranking Methods}\label{sec:app:reranking}
\paragraph{\textbf{Query-Time Score Regularization}}
Based on the cluster hypothesis in IR, \textit{query-time score regularization}~\cite{diaz2005regularizing-adhoc, diaz2007regularizing-querybased} adjusts retrieval scores so that semantically similar documents receive similar relevance scores. 
Score regularization can be seen as a generalization of pseudo-relevance feedback (PRF), a well-established method for improving retrieval performance \cite{rocchio1971relevance, wang2023colbert}. However, unlike PRF, which modifies query representations, score regularization directly refines retrieval scores. Although shown to improve retrieval performance, it remains underexplored in dense retrieval and especially in a RAG setup.

Reranking procedure is as follows:
\begin{enumerate}
  \item \textbf{Initial Retrieval:} Pinecone is queried, returning $k$ documents and their $n$-dim embeddings, $\{d_i\}_{i=1}^k$.
  \item \textbf{Similarity Matrix:} From a  matrix $D \in \mathbb{R}^{k \times n}$, where each row is $d_i$, we compute a similarity matrix $W = D D^T \in \mathbb{R}^{k \times k}$.
  \item \textbf{Row-Stochastic Matrix:} For each row of $W$, we keep top-$m$ similarities, normalize to sum to 1, yielding a row-stochastic matrix $P$. 
  \item \textbf{Regularized Scores:} Given an original score vector $s\in \mathbb{R}^{k \times 1}$, we define $\tilde{s} = P^t s$,
  where $t$ controls the strength. We then rerank documents by $\tilde{s}$, enhancing clusters of semantically similar passages.
\end{enumerate}

\paragraph{\textbf{Stochastic Reranking}}
\textit{Stochastic retrieval} uses Plackett-Luce sampling \cite{oosterhuis2021PLRank} on the initial score distribution to promote item-fairness \cite{diaz_2020_stochastic_ranking}. As shown by~\citet{kim2025fairragimpactfair}, this can boost item-fairness in RAG systems and potentially enhance downstream QA performance. In our reranking, we randomly select one ranking from the 50 sampled rankings, where sampling intensity parameter is set to 2.

\section{Falcon Prompts}\label{sec:app:prompts}
\begin{itemize}
    \item \{INFO\_PROMPT\} is first constructed by iterating through the retrieved documents by filling in the following template:\\
    \texttt{Document \{DOC\_i\}: \{TEXT\}}.
    \item Fallback prompt is selected when the generator produces an ill-formatted output (e.g., no \texttt{<answer>} tag exists).
    \item Non-RAG prompt is selected when the router selects $\rModel_0$ and no external information is retrieved.
\end{itemize}

\subsection{RAG Reasoning Prompt}
\begin{tcolorbox}
You must answer a user question, based on the information (web documents) provided.
Before answering the question, you must conduct your reasoning inside <think> and </think>. During this reasoning step, think about the intent of the user's question and focus on evaluating the relevance and helpfulness of each document to the question. This is important because the information comes from web retrieval and may include irrelevant content. By explicitly reasoning through the relevance and utility of each document, you should seek to ensure that your final answer is accurate and grounded in the pertinent information. After that, think step by step to get to the right answer.
Generation format: you need to surround your reasoning with <think> and </think>, and need to surround your answer with <answer> and </answer>.
For example: <think> your reasoning </think>; <answer> your answer </answer>.

User Question: \{QUESTION\}

Information:
\{INFO\_PROMPT\}

Show your reasoning between <think> and </think>, and provide your final answer between <answer> and </answer>.
\end{tcolorbox}

\subsection{RAG Fallback Prompt}
\begin{tcolorbox}
Using the information provided below, please answer the user's question. Consider the intent of the user's question and focus on the most relevant information that directly addresses what they're asking. Make sure your response is accurate and based on the provided information.

Question: \{QUESTION\}

Information:
\{INFO\_PROMPT\}

Provide a clear and direct answer to "\{QUESTION\}" based on the information above.
\end{tcolorbox}

\subsection{Non-RAG Reasoning Prompt}
\begin{tcolorbox}
You must answer the given user question enclosed within <question> and </question>.
Before answering the question, you must conduct your reasoning inside <think> and </think>. During this reasoning step, think about the intent of the user's question then think step by step to get to the right answer.
Generation format: you need to surround your reasoning with <think> and </think>, and need to surround your answer with <answer> and </answer>.
For example: <think> your reasoning </think>; <answer> your answer </answer>.

User Question:
\{QUESTION\}

Show your reasoning between <think> and </think>, and provide your final answer between <answer> and </answer>.
\end{tcolorbox}

\subsection{Non-RAG Fallback Prompt}
\begin{tcolorbox}
Answer the following question with ONLY the answer. No explanations, reasoning, or additional context.

Question: \{QUESTION\}
\end{tcolorbox}

\section{LTRR Loss Functions}\label{sec:app:ltrr-loss}
\paragraph{Pointwise Approach} Each retriever's utility gain $\utilityGain_i$ is independently predicted by minimizing a regression loss:
\begin{align}
    L_{point}(\ltrrScorer) = \sum_{\query\in\querySpace}\sum_i (\utilityGain_i - \ltrrScorer(\features(\query, \rModel_i)))^2  .
\end{align}

\paragraph{Pairwise Approach} The model learns from pairwise comparisons between retrievers, minimizing ranking inversions through a pairwise loss, where ranking preference ($\rModel_i \succ_\query \rModel_j$) is determined based on the relative utility gain each retriever provides to the RAG system compared to the no-retrieval baseline:
\begin{align}
    L_{pair}(\ltrrScorer) = \sum_{\query\in\querySpace} \sum_{\rModel_i \succ_\query \rModel_j} \slIndicator{\ltrrScorer(\features(\query, \rModel_i)) < \ltrrScorer(\features(\query, \rModel_j))}  .
\end{align}

\paragraph{Listwise Approach} The model directly optimizes the predicted utility gains across all retrievers for each query. With $P_i$ and $Q_i$ being the probability distribution over rankings based on utility-gain and model prediction, we define the listwise loss as:
\begin{align}
    L_{list}(\ltrrScorer) = \sum_{\query\in\querySpace}\sum_i -P_i(\utilityGain_i) log(Q_i(\ltrrScorer(\features(\query, \rModel_i))))  .
\end{align}

\newpage
\section{Dataset Generation via DataMorgana}\label{sec:app:dm-tables}
Table \ref{tab:dm-question} describes question type configurations, and Table \ref{tab:dm-user} describes user type configuration for dataset generation using DataMorgana \cite{filice2025datamorgana}.
\input{99-appendix/dm-question-table}
\input{99-appendix/dm-user-table}

\section{Data Statistics}\label{sec:app:data-stat}
Table \ref{tab:answer-type-stats} shows statistics of training and test data from the DataMorgana-generated dataset.
\input{99-appendix/data-stat}

\section{Standard RAG Performance}\label{sec:app:single-ret}
Table \ref{tab:single-ret-results} shows the average downstream utilities of standard RAG models.
Models with the highest utility value were selected as the `Best Standard RAG' models in Table \ref{tab:main-results}.
\input{99-appendix/single-retriever}

%% file: 99-appendix/dm-question-table.tex
\begin{table}[ht]
\centering
\renewcommand{\arraystretch}{1.4}
\begin{tabularx}{\textwidth}{|c|c|c|X|}
\hline
\textbf{Categorization} & \textbf{Category} & \textbf{Prob} &\textbf{Description} \\
\hline
\multirow{2}{*}{Answer Type} 
    & factoid & 0.4 & a question seeking a specific, concise piece of information or a fact about a particular subject. \\
    & multi-aspect & 0.2 & A question about two different aspects of the same entity/concept. For example: `What are the advantages of AI-powered diagnostics, and what are the associated risks of bias in medical decision-making?', `How do cryptocurrencies enable financial inclusion, and what are the security risks associated with them?'. The information required to answer the question needs to come from two documents, specifically, the first document must provide information about the first aspect, while the second must provide information about the second aspect. \\
    & comparison & 0.2 & a comparison question that requires comparing two related concepts or entities. The comparison must be natural and reasonable, i.e., comparing two entities by a common attribute which is meaningful and relevant to both entities. For example: 'Who is older, Glenn Hughes or Ross Lynch?', `Are Pizhou and Jiujiang in the same province?', `Pyotr Ilyich Tchaikovsky and Giuseppe Verdi have this profession in common'. The information required to answer the question needs to come from two documents, specifically, the first document must provide information about the first entity/concept, while the second must provide information about the second entity/concept. \\
    & open-ended & 0.2 & a question seeking a detailed or exploratory response, encouraging discussion or elaboration. \\
\hline
\multirow{2}{*}{Premise} 
    & without-premise & 0.7 & a question that does not contain any premise or any information about the user. \\
    & with-premise & 0.3 & a question starting with a very short premise, where the user reveals one's needs or some information about himself. \\
\hline
\multirow{2}{*}{Phrasing} 
    & concise-and-natural & 0.25 & a concise, direct, and natural question consisting of a few words. \\
    & verbose-and-natural & 0.25 & a relatively long question consisting of more than 9 words. \\
    & short-search-query & 0.25 & a question phrased as a typed web query for search engines (only keywords, without punctuation and without a natural-sounding structure). It consists of less than 7 words. \\
    & long-search-query & 0.25 & a question phrased as a typed web query for search engines (only keywords without punctuation and without a natural-sounding structure). It consists of more than 6 words. \\
\hline
\multirow{2}{*}{Linguistic Variation}
    & similar-to-document & 0.5 & a question that is written using the same or similar terminology and phrases appearing in the documents. \\
    & distant-from-document & 0.5 & a question that is written using the terms completely different from the ones appearing in the documents. \\
\hline
\end{tabularx}
\caption{Question categorizations and descriptions.}
\label{tab:dm-question}
\vspace{-10pt}
\end{table}

%% file: 99-appendix/dm-user-table.tex
\begin{table}[ht]
\centering
\renewcommand{\arraystretch}{1.4}
\begin{tabularx}{\textwidth}{|c|c|c|X|}
\hline
\textbf{Categorization} & \textbf{Category} & \textbf{Prob} &\textbf{Description} \\
\hline
\multirow{2}{*}{User Expertise} 
    & expert & 0.4 & an expert on the subject discussed in the documents, therefore he asks complex questions. \\
    & novice & 0.6 & a person with basic knowledge on the topic discussed in the documents, therefore, he asks non-complex questions. \\
\hline
\end{tabularx}
\caption{User categorization and descriptions.}
\label{tab:dm-user}
\end{table}

%% file: 99-appendix/data-stat.tex
\begin{table}[ht]
\centering
\begin{tabular}{lcc}
\toprule
 & \textbf{Train} & \textbf{Test} \\
\midrule
Balanced       & 7,995 & 1999 \\
multi-aspect   & 6442 & 410 \\
comparison     & 6457 & 385 \\
complex        & 4751 & 787 \\
open-ended     & 6345 & 411 \\
\bottomrule
\end{tabular}
\caption{The number of queries in each dataset for training and testing LTRR algorithms.}
\label{tab:answer-type-stats}
\end{table}

%% file: 99-appendix/single-retriever.tex
\begin{table}[ht]
\centering
\resizebox{\textwidth}{!}{
\begin{tabular}{c ccccc c ccccc}
   & \multicolumn{5}{c}{BEM-based}   && \multicolumn{5}{c}{AC-based} \\
   \cline{2-6} \cline{8-12}
   Model & \makecell{Balanced} & \makecell{Unseen\\multi-aspect}   & \makecell{Unseen\\comparison} & \makecell{Unseen\\complex} & \makecell{Unseen\\open-ended} && \makecell{Balanced} & \makecell{Unseen\\multi-aspect}   & \makecell{Unseen\\comparison} & \makecell{Unseen\\complex} & \makecell{Unseen\\open-ended} \\
   \hline
BM25            & \textbf{0.3599} & 0.1826 & 0.3034  & \textbf{0.3595}  & 0.3430  && 0.5769 & 0.5656 & 0.5706 & \textbf{0.5812} & 0.5711 \\
BM25+Stochastic & 0.3302 & 0.1796 & 0.3046  & 0.3252  & 0.3042  && 0.5510 & 0.5688 & 0.5469 & 0.5543 & 0.5424 \\
BM25+Regularize & 0.3539 & 0.1893 & 0.3061  & 0.3559  & 0.3350  && \textbf{0.5776} & 0.5696 & \textbf{0.5723} & 0.5790 & 0.5749 \\
E5              & 0.3538 & \textbf{0.1941} & \textbf{0.3074}  & 0.3496  & \textbf{0.3465}  && 0.5769 & 0.5791 & 0.5711 & 0.5782 & \textbf{0.5784} \\
E5+Stochastic   & 0.3258 & 0.1855 & 0.3049  & 0.3252  & 0.3002  && 0.5503 & 0.5805 & 0.5483 & 0.5417 & 0.5546 \\
E5+Regularize   & 0.3418 & 0.1829 & 0.2991  & 0.3490  & 0.3083  && 0.5694 & \textbf{0.5809} & 0.5692 & 0.5756 & 0.5620 \\

\end{tabular}
}
\caption{Average downstream utility measured by either BEM or AC of standard RAG systems with a single retriever model. Boldface indicates the highest average downstream utility value.
}
\label{tab:single-ret-results}
\end{table}

%% file: XX-references.bib
@article{lewis2020retrieval,
  title={Retrieval-augmented generation for knowledge-intensive nlp tasks},
  author={Lewis, Patrick and Perez, Ethan and Piktus, Aleksandra and Petroni, Fabio and Karpukhin, Vladimir and Goyal, Naman and K{\"u}ttler, Heinrich and Lewis, Mike and Yih, Wen-tau and Rockt{\"a}schel, Tim and others},
  journal={Advances in neural information processing systems},
  volume={33},
  pages={9459--9474},
  year={2020}
}

@inproceedings{kim2025fairragimpactfair, author = {Kim, To Eun and Diaz, Fernando}, title = {Towards Fair RAG: On the Impact of Fair Ranking in Retrieval-Augmented Generation}, year = {2025}, isbn = {9798400718618}, publisher = {Association for Computing Machinery}, address = {New York, NY, USA}, url = {https://doi.org/10.1145/3731120.3744599}, doi = {10.1145/3731120.3744599}, booktitle = {Proceedings of the 2025 International ACM SIGIR Conference on Innovative Concepts and Theories in Information Retrieval (ICTIR)}, pages = {33–43}, numpages = {11}, location = {Padua, Italy}, series = {ICTIR '25} }

@article{kim2024reml,
  title={Retrieval-Enhanced Machine Learning: Synthesis and Opportunities},
  author={Kim, To Eun and Salemi, Alireza and Drozdov, Andrew and Diaz, Fernando and Zamani, Hamed},
  journal={arXiv preprint arXiv:2407.12982},
  year={2024}
}

@inproceedings{oosterhuis2021PLRank,
  title={Computationally efficient optimization of plackett-luce ranking models for relevance and fairness},
  author={Oosterhuis, Harrie},
  booktitle={Proceedings of the 44th International ACM SIGIR Conference on Research and Development in Information Retrieval},
  pages={1023--1032},
  year={2021}
}

@inproceedings{diaz_2020_stochastic_ranking,
	series = {{CIKM} '20},
	title = {Evaluating {Stochastic} {Rankings} with {Expected} {Exposure}},
	booktitle = {Proceedings of the 29th {ACM} {International} {Conference} on {Information} \& {Knowledge} {Management}},
	publisher = {Association for Computing Machinery},
	author = {Diaz, Fernando and Mitra, Bhaskar and Ekstrand, Michael D. and Biega, Asia J. and Carterette, Ben},
	year = {2020},
	pages = {275--284}
}

@misc{Falcon3,
    title = {The Falcon 3 Family of Open Models},
    url = {https://huggingface.co/blog/falcon3},
    author = {Falcon-LLM Team},
    month = {December},
    year = {2024}
}

@article{robertson2009probabilistic,
  title={The probabilistic relevance framework: BM25 and beyond},
  author={Robertson, Stephen and Zaragoza, Hugo and others},
  journal={Foundations and trends{\textregistered} in information retrieval},
  volume={3},
  number={4},
  pages={333--389},
  year={2009},
  publisher={Now Publishers, Inc.}
}

@inproceedings{zamani:reml,
	author = {Hamed Zamani and Fernando Diaz and Mostafa Dehghani and Donald Metzler and Michael Bendersky},
	booktitle = {Proceedings of the 45th Annual International ACM SIGIR Conference on Research and Development in Information Retrieval},
	title = {Retrieval-Enhanced Machine Learning},
	year = {2022}}

@incollection{callan2002distributed,
  title={Distributed information retrieval},
  author={Callan, Jamie},
  booktitle={Advances in information retrieval: recent research from the center for intelligent information retrieval},
  pages={127--150},
  year={2002},
  publisher={Springer}
}

@article{chen2001metaspider,
  title={MetaSpider: Meta-searching and categorization on the Web},
  author={Chen, Hsinchun and Fan, Haiyan and Chau, Michael and Zeng, Daniel},
  journal={Journal of the American Society for Information Science and Technology},
  volume={52},
  number={13},
  pages={1134--1147},
  year={2001},
  publisher={Wiley Online Library}
}

@inproceedings{callan1995searching,
  title={Searching distributed collections with inference networks},
  author={Callan, James P and Lu, Zhihong and Croft, W Bruce},
  booktitle={Proceedings of the 18th annual international ACM SIGIR conference on Research and development in information retrieval},
  pages={21--28},
  year={1995}
}

@article{diaz2007regularizing-querybased,
  title={Regularizing query-based retrieval scores},
  author={Diaz, Fernando},
  journal={Information Retrieval},
  volume={10},
  pages={531--562},
  year={2007},
  publisher={Springer}
}

@inproceedings{Khramtsova23whichRetriever, author = {Khramtsova, Ekaterina and Zhuang, Shengyao and Baktashmotlagh, Mahsa and Wang, Xi and Zuccon, Guido}, title = {Selecting which Dense Retriever to use for Zero-Shot Search}, year = {2023}, isbn = {9798400704086}, publisher = {Association for Computing Machinery}, url = {https://doi.org/10.1145/3624918.3625330}, doi = {10.1145/3624918.3625330}, booktitle = {Proceedings of the Annual International ACM SIGIR Conference on Research and Development in Information Retrieval in the Asia Pacific Region}, pages = {223–233}, numpages = {11}, location = {Beijing, China}, series = {SIGIR-AP '23} }

@inproceedings{Khramtsova24RetrieverRanking, author = {Khramtsova, Ekaterina and Zhuang, Shengyao and Baktashmotlagh, Mahsa and Zuccon, Guido}, title = {Leveraging LLMs for Unsupervised Dense Retriever Ranking}, year = {2024}, isbn = {9798400704314}, publisher = {Association for Computing Machinery}, url = {https://doi.org/10.1145/3626772.3657798}, doi = {10.1145/3626772.3657798}, booktitle = {Proceedings of the 47th International ACM SIGIR Conference on Research and Development in Information Retrieval}, pages = {1307–1317}, numpages = {11}, location = {Washington DC, USA}, series = {SIGIR '24} }

@inproceedings{diaz2007performance,
  title={Performance prediction using spatial autocorrelation},
  author={Diaz, Fernando},
  booktitle={Proceedings of the 30th annual international ACM SIGIR conference on Research and development in information retrieval},
  pages={583--590},
  year={2007}
}

@inproceedings{diaz2005regularizing-adhoc,
  title={Regularizing ad hoc retrieval scores},
  author={Diaz, Fernando},
  booktitle={Proceedings of the 14th ACM international conference on Information and knowledge management},
  pages={672--679},
  year={2005}
}

@article{rocchio1971relevance,
  title={Relevance feedback in information retrieval},
  author={Rocchio Jr, Joseph John},
  journal={The SMART retrieval system: experiments in automatic document processing},
  year={1971},
  publisher={Englewood Cliffs}
}

@article{wang2023colbert,
  title={ColBERT-PRF: Semantic pseudo-relevance feedback for dense passage and document retrieval},
  author={Wang, Xiao and Macdonald, Craig and Tonellotto, Nicola and Ounis, Iadh},
  journal={ACM Transactions on the Web},
  volume={17},
  number={1},
  pages={1--39},
  year={2023},
  publisher={ACM New York, NY}
}

@inproceedings{filice2025datamorgana,
  title={Generating Q\&A benchmarks for RAG evaluation in enterprise settings},
  author={Filice, Simone and Horowitz, Guy and Carmel, David and Karnin, Zohar and Lewin-Eytan, Liane and Maarek, Yoelle},
  booktitle={Proceedings of the 63rd Annual Meeting of the Association for Computational Linguistics (Volume 6: Industry Track)},
  pages={469--484},
  year={2025}
}

@article{penedo2025fineweb,
  title={The fineweb datasets: Decanting the web for the finest text data at scale},
  author={Penedo, Guilherme and Kydl{\'\i}{\v{c}}ek, Hynek and Lozhkov, Anton and Mitchell, Margaret and Raffel, Colin A and Von Werra, Leandro and Wolf, Thomas and others},
  journal={Advances in Neural Information Processing Systems},
  volume={37},
  pages={30811--30849},
  year={2025}
}

@inproceedings{diaz2010federatedAggregated,
  title={From federated to aggregated search},
  author={Diaz, Fernando and Lalmas, Mounia and Shokouhi, Milad},
  booktitle={Proceedings of the 33rd international ACM SIGIR conference on Research and development in information retrieval},
  pages={910--910},
  year={2010}
}

@article{arguello2017aggregated,
  title={Aggregated search},
  author={Arguello, Jaime and others},
  journal={Foundations and Trends{\textregistered} in Information Retrieval},
  volume={10},
  number={5},
  pages={365--502},
  year={2017},
  publisher={Now Publishers, Inc.}
}

@inproceedings{glover1999metasearch,
  title={Architecture of a metasearch engine that supports user information needs},
  author={Glover, Eric J and Lawrence, Steve and Birmingham, William P and Giles, C Lee},
  booktitle={Proceedings of the eighth international conference on Information and knowledge management},
  pages={210--216},
  year={1999}
}

@inproceedings{mu-etal-2024-query,
    title = "Query Routing for Homogeneous Tools: An Instantiation in the {RAG} Scenario",
    author = "Mu, Feiteng  and
      Jiang, Yong  and
      Zhang, Liwen  and
      Liuchu, Liuchu  and
      Li, Wenjie  and
      Xie, Pengjun  and
      Huang, Fei",
    booktitle = "Findings of the Association for Computational Linguistics: EMNLP 2024",
    month = nov,
    year = "2024",
    address = "Miami, Florida, USA",
    publisher = "Association for Computational Linguistics",
    url = "https://aclanthology.org/2024.findings-emnlp.598/",
    doi = "10.18653/v1/2024.findings-emnlp.598",
    pages = "10225--10230"
}

@inproceedings{jeong-etal-2024-adaptiveRAG,
    title = "Adaptive-{RAG}: Learning to Adapt Retrieval-Augmented Large Language Models through Question Complexity",
    author = "Jeong, Soyeong  and
      Baek, Jinheon  and
      Cho, Sukmin  and
      Hwang, Sung Ju  and
      Park, Jong",
    editor = "Duh, Kevin  and
      Gomez, Helena  and
      Bethard, Steven",
    booktitle = "Proceedings of the 2024 Conference of the North American Chapter of the Association for Computational Linguistics: Human Language Technologies (Volume 1: Long Papers)",
    month = jun,
    year = "2024",
    address = "Mexico City, Mexico",
    publisher = "Association for Computational Linguistics",
    url = "https://aclanthology.org/2024.naacl-long.389/",
    doi = "10.18653/v1/2024.naacl-long.389",
    pages = "7036--7050"
}

@inproceedings{Guerraoui25FederatedRAG, author = {Guerraoui, Rachid and Kermarrec, Anne-Marie and Petrescu, Diana and Pires, Rafael and Randl, Mathis and de Vos, Martijn}, title = {Efficient Federated Search for Retrieval-Augmented Generation}, year = {2025}, isbn = {9798400715389}, publisher = {Association for Computing Machinery}, url = {https://doi.org/10.1145/3721146.3721942}, doi = {10.1145/3721146.3721942}, booktitle = {Proceedings of the 5th Workshop on Machine Learning and Systems}, pages = {74–81}, numpages = {8}, location = {World Trade Center, Rotterdam, Netherlands}, series = {EuroMLSys '25} 
}

@inproceedings{tang2024adapting, author = {Tang, Xiaqiang and Li, Jian and Du, Nan and Xie, Sihong}, title = {Adapting to non-stationary environments: multi-armed bandit enhanced retrieval-augmented generation on knowledge graphs}, year = {2025}, isbn = {978-1-57735-897-8}, publisher = {AAAI Press}, url = {https://doi.org/10.1609/aaai.v39i12.33380}, doi = {10.1609/aaai.v39i12.33380}, booktitle = {Proceedings of the Thirty-Ninth AAAI Conference on Artificial Intelligence and Thirty-Seventh Conference on Innovative Applications of Artificial Intelligence and Fifteenth Symposium on Educational Advances in Artificial Intelligence}, articleno = {1407}, numpages = {9}, series = {AAAI'25/IAAI'25/EAAI'25} }

@inproceedings{dai2017ltrResources,
  title={Learning to rank resources},
  author={Dai, Zhuyun and Kim, Yubin and Callan, Jamie},
  booktitle={Proceedings of the 40th International ACM SIGIR conference on research and development in information retrieval},
  pages={837--840},
  year={2017}
}

@inproceedings{salemi2024erag,
  title={Evaluating retrieval quality in retrieval-augmented generation},
  author={Salemi, Alireza and Zamani, Hamed},
  booktitle={Proceedings of the 47th International ACM SIGIR Conference on Research and Development in Information Retrieval},
  pages={2395--2400},
  year={2024}
}

@inproceedings{
jin2025searchR1,
title={Search-R1: Training {LLM}s to Reason and Leverage Search Engines with Reinforcement Learning},
author={Bowen Jin and Hansi Zeng and Zhenrui Yue and Jinsung Yoon and Sercan O Arik and Dong Wang and Hamed Zamani and Jiawei Han},
booktitle={Second Conference on Language Modeling},
year={2025},
url={https://openreview.net/forum?id=Rwhi91ideu}
}

@inproceedings{chen2025ReSearch,
  title={Learning to Reason with Search for LLMs via Reinforcement Learning},
  author={Chen, Mingyang and Sun, Linzhuang and Li, Tianpeng and Zhu, Chenzheng and Wang, Haofen and Pan, Jeff Z and Zhang, Wen and Chen, Huajun and Yang, Fan and Zhou, Zenan and others},
  booktitle={The Thirty-ninth Annual Conference on Neural Information Processing Systems}
}

@article{luo2025evaluation,
  title={Evaluation Report on MCP Servers},
  author={Luo, Zhiling and Shi, Xiaorong and Lin, Xuanrui and Gao, Jinyang},
  journal={arXiv preprint arXiv:2504.11094},
  year={2025}
}

@misc{anthropic2024mcp,
  author       = {Anthropic},
  title        = {Introducing the Model Context Protocol},
  year         = {2024},
  month        = {November},
  url          = {https://www.anthropic.com/news/model-context-protocol},
  note         = {Accessed: 2025-05-23}
}

@inproceedings{bulian2022-tomayto-bem,
    title = "Tomayto, Tomahto. Beyond Token-level Answer Equivalence for Question Answering Evaluation",
    author = {Bulian, Jannis  and
      Buck, Christian  and
      Gajewski, Wojciech  and
      B{\"o}rschinger, Benjamin  and
      Schuster, Tal},
    editor = "Goldberg, Yoav  and
      Kozareva, Zornitsa  and
      Zhang, Yue",
    booktitle = "Proceedings of the 2022 Conference on Empirical Methods in Natural Language Processing",
    month = dec,
    year = "2022",
    address = "Abu Dhabi, United Arab Emirates",
    publisher = "Association for Computational Linguistics",
    url = "https://aclanthology.org/2022.emnlp-main.20/",
    doi = "10.18653/v1/2022.emnlp-main.20",
    pages = "291--305"
}

@inproceedings{es-etal-2024-ragas,
    title = "{RAGA}s: Automated Evaluation of Retrieval Augmented Generation",
    author = "Es, Shahul  and
      James, Jithin  and
      Espinosa Anke, Luis  and
      Schockaert, Steven",
    editor = "Aletras, Nikolaos  and
      De Clercq, Orphee",
    booktitle = "Proceedings of the 18th Conference of the European Chapter of the Association for Computational Linguistics: System Demonstrations",
    month = mar,
    year = "2024",
    address = "St. Julians, Malta",
    publisher = "Association for Computational Linguistics",
    url = "https://aclanthology.org/2024.eacl-demo.16",
    pages = "150--158"
}

@article{oro2024evaluating,
  title={Evaluating Retrieval-Augmented Generation for Question Answering with Large Language Models},
  author={Oro, Ermelinda and Granata, Francesco Maria and Lanza, Antonio and Bachir, Amir and De Grandis, Luca and Ruffolo, Massimo},
  year={2024}
}

@inproceedings{mallen-etal-2023-trust,
    title = "When Not to Trust Language Models: Investigating Effectiveness of Parametric and Non-Parametric Memories",
    author = "Mallen, Alex  and
      Asai, Akari  and
      Zhong, Victor  and
      Das, Rajarshi  and
      Khashabi, Daniel  and
      Hajishirzi, Hannaneh",
    editor = "Rogers, Anna  and
      Boyd-Graber, Jordan  and
      Okazaki, Naoaki",
    booktitle = "Proceedings of the 61st Annual Meeting of the Association for Computational Linguistics (Volume 1: Long Papers)",
    month = jul,
    year = "2023",
    address = "Toronto, Canada",
    publisher = "Association for Computational Linguistics",
    url = "https://aclanthology.org/2023.acl-long.546/",
    doi = "10.18653/v1/2023.acl-long.546",
    pages = "9802--9822"
}

@inproceedings{cao2007ltr,
  title={Learning to rank: from pairwise approach to listwise approach},
  author={Cao, Zhe and Qin, Tao and Liu, Tie-Yan and Tsai, Ming-Feng and Li, Hang},
  booktitle={Proceedings of the 24th international conference on Machine learning},
  pages={129--136},
  year={2007}
}

@inproceedings{lee2024routerretriever, author = {Lee, Hyunji and Soldaini, Luca and Cohan, Arman and Seo, Minjoon and Lo, Kyle}, title = {ROUTERRETRIEVER: routing over a mixture of expert embedding models}, year = {2025}, isbn = {978-1-57735-897-8}, publisher = {AAAI Press}, url = {https://doi.org/10.1609/aaai.v39i11.33306}, doi = {10.1609/aaai.v39i11.33306}, booktitle = {Proceedings of the Thirty-Ninth AAAI Conference on Artificial Intelligence and Thirty-Seventh Conference on Innovative Applications of Artificial Intelligence and Fifteenth Symposium on Educational Advances in Artificial Intelligence}, articleno = {1333}, numpages = {9}, series = {AAAI'25/IAAI'25/EAAI'25} }

@inproceedings{chen2016xgboost,
  title={Xgboost: A scalable tree boosting system},
  author={Chen, Tianqi and Guestrin, Carlos},
  booktitle={Proceedings of the 22nd acm sigkdd international conference on knowledge discovery and data mining},
  pages={785--794},
  year={2016}
}

@inproceedings{joachims2006training,
  title={Training linear SVMs in linear time},
  author={Joachims, Thorsten},
  booktitle={Proceedings of the 12th ACM SIGKDD international conference on Knowledge discovery and data mining},
  pages={217--226},
  year={2006}
}

@article{he2021debertav3,
  title={Debertav3: Improving deberta using electra-style pre-training with gradient-disentangled embedding sharing},
  author={He, Pengcheng and Gao, Jianfeng and Chen, Weizhu},
  journal={arXiv preprint arXiv:2111.09543},
  year={2021}
}

@article{wu2010adapting,
  title={Adapting boosting for information retrieval measures},
  author={Wu, Qiang and Burges, Christopher JC and Svore, Krysta M and Gao, Jianfeng},
  journal={Information Retrieval},
  volume={13},
  pages={254--270},
  year={2010},
  publisher={Springer}
}

@inproceedings{cormack2009reciprocal,
  author =        {Cormack, Gordon V and Clarke, Charles LA and
                   Buettcher, Stefan},
  booktitle =     {Proceedings of the 32nd international ACM SIGIR
                   conference on Research and development in information
                   retrieval},
  pages =         {758--759},
  title =         {Reciprocal rank fusion outperforms condorcet and
                   individual rank learning methods},
  year =          {2009},
}

@article{wang2022text,
  title={Text Embeddings by Weakly-Supervised Contrastive Pre-training},
  author={Wang, Liang and Yang, Nan and Huang, Xiaolong and Jiao, Binxing and Yang, Linjun and Jiang, Daxin and Majumder, Rangan and Wei, Furu},
  journal={arXiv preprint arXiv:2212.03533},
  year={2024}
}

@article{liu2009learning,
  title={Learning to rank for information retrieval},
  author={Liu, Tie-Yan and others},
  journal={Foundations and Trends{\textregistered} in Information Retrieval},
  volume={3},
  number={3},
  pages={225--331},
  year={2009},
  publisher={Now Publishers, Inc.}
}

@inproceedings{kalra-etal-2025-mor,
    title = "{M}o{R}: Better Handling Diverse Queries with a Mixture of Sparse, Dense, and Human Retrievers",
    author = "Kalra, Jushaan Singh  and
      Zhao, Xinran  and
      Kim, To Eun  and
      Cai, Fengyu  and
      Diaz, Fernando  and
      Wu, Tongshuang",
    editor = "Christodoulopoulos, Christos  and
      Chakraborty, Tanmoy  and
      Rose, Carolyn  and
      Peng, Violet",
    booktitle = "Proceedings of the 2025 Conference on Empirical Methods in Natural Language Processing",
    month = nov,
    year = "2025",
    address = "Suzhou, China",
    publisher = "Association for Computational Linguistics",
    url = "https://aclanthology.org/2025.emnlp-main.601/",
    doi = "10.18653/v1/2025.emnlp-main.601",
    pages = "11971--11990",
    ISBN = "979-8-89176-332-6"
}
